\documentclass[letterpaper, 10 pt, conference]{ieeeconf}  

\IEEEoverridecommandlockouts                              
\usepackage[utf8]{inputenc}
\usepackage[T1]{fontenc}
\usepackage{graphicx}
\usepackage{booktabs}

\title{\LARGE \bf
Structured Insight from Unstructured Data: Large Language Models for SDOH-Driven Diabetes Risk Prediction
}

\author{Sasha Ronaghi$^{1}$, Prerit Choudhary$^{1}$, David H. Rehkopf$^{2}$, and Bryant Lin$^{3}$
\thanks{*This work was supported by the Stanford Center for Asian Health Research and Education (CARE).
}
\thanks{$^{1}$Sasha Ronaghi, Prerit Choudhary are with the Department of Computer Science, Stanford University, Stanford, CA 94305 USA
        {\tt\small sronaghi@stanford.edu, preritc@stanford.edu}}%
\thanks{$^{2}$David H. Rekhopf is with the Department of Epidemiology and Population Health, Department of Medicine (Division of Primary Care and Population Health), Department of Pediatrics, and Department of Health Policy, Stanford University School of Medicine, and Department of Sociology, Stanford University, Stanford, CA  94305.
        {\tt\small drehkopf@stanford.edu}}%
\thanks{$^{3}$Bryant Lin is with the Department of Medicine, Stanford University School of Medicine, Stanford, CA 94305 USA.
        {\tt\small 
bylin@stanford.edu}}%
}

\begin{document}

\maketitle
\thispagestyle{empty}
\pagestyle{empty}

\begin{abstract}

Social determinants of health (SDOH) play a critical role in Type 2 Diabetes (T2D) management but are often absent from electronic health records and risk prediction models. Most individual-level SDOH data is collected through structured screening tools, which lack the flexibility to capture the complexity of patient experiences and unique needs of a clinic’s population. This study explores the use of large language models (LLMs) to extract structured SDOH information from unstructured patient life stories and evaluate the predictive value of both the extracted features and the narratives themselves for assessing diabetes control. We collected unstructured interviews from 65 T2D patients aged 65 and older, focused on their lived experiences, social context, and diabetes management. These narratives were analyzed using LLMs with retrieval-augmented generation to produce concise, actionable qualitative summaries for clinical interpretation and structured quantitative SDOH ratings for risk prediction modeling. The structured SDOH ratings were used independently and in combination with traditional laboratory biomarkers as inputs to linear and tree-based machine learning models (Ridge, Lasso, Random Forest, and XGBoost) to demonstrate how unstructured narrative data can be applied in conventional risk prediction workflows. Finally, we evaluated several LLMs on their ability to predict a patient’s level of diabetes control (low, medium, high) directly from interview text with A1C values redacted. LLMs achieved 60\% accuracy in predicting diabetes control levels from interview text. This work demonstrates how LLMs can translate unstructured SDOH-related data into structured insights, offering a scalable approach to augment clinical risk models and decision-making.
\end{abstract}

\section{INTRODUCTION}
Social determinants of health (SDOH) such as financial stress, access to transportation, and social support—play a significant role in shaping diabetes experience and outcomes [1]. However, social determinants are often missing from electronic health record (EHR) data, which serves as the foundation for most risk prediction models and provides essential context for clinicians delivering care [2, 3]. Truong et al show that only 1.9\% of admissions include ICD-10 Z-codes, which are the only diagnosis codes that identify non-medical factors that may affect a patient's health [4].

Traditionally, questionnaire-style screenings have been the primary method for collecting individual-level SDOH data [5] as they allow for large-scale data collection and potential integration  with EHR models. Commonly used tools include PRAPARE [6] and the Center for Medicare and Medicaid Services’ Accountable Health Communities screening tool [7]. The limited structure and pre-defined categories within these tools often fail to capture the richness of patients' lived experiences and the unique needs and circumstances of different patient demographics. To overcome this challenge, clinics are increasingly utilizing home-grown, customized questionnaires tailored to a clinic’s patient demographics, which can present data interoperability challenges for larger-scale population health analysis [5]. This trend highlights the need for a more flexible mode of SDOH data collection. Additionally, although diabetes is influenced by behaviors, psychological
characteristics, and socioeconomic factors [8], existing risk prediction models do not integrate these domains [9]. This underscores the need for larger-scale data collection methodologies that capture behaviors, psychological characteristics, and socioeconomic factors in order to develop more holistic diabetes risk prediction tools.

Collecting social determinant data in an unstructured manner, such as through clinical encounters or free-text boxes, presents an opportunity for capturing richer, more nuanced details about patients’ lives, including aspects of their disease management and social context that are rarely addressed in structured data collection methods [10]. However, until recently, the analysis of unstructured data using traditional machine learning models was constrained by the need for extensive manual preprocessing and feature extraction [11]. Large language models (LLMs) now present a transformative opportunity to process and analyze unstructured language efficiently and at scale [12]. LLMs excel at understanding context, identifying patterns, and extracting key insights from free-text narratives, enabling the discovery of novel risk factors and relationships that were previously inaccessible. 

This study leverages the capabilities of LLMs to analyze unstructured life stories from 65 T2D patients aged 65 years and older, focusing on their day-to-day disease management, diabetes literacy, and SDOH-related challenges. Although we apply these tools to a relatively small sample size, this work is aimed to serve as a proof of concept to illustrate how LLMs can analyze large quantities of unstructured qualitative data quickly for use within the clinical setting and for quantitative risk prediction modeling. 

Importantly, the study was conducted in a cohort with high Asian American and Pacific Islanders (AAPI) representation (46.2\% of participants), providing valuable insight into this population. AAPI have a higher type 2 diabetes (T2D) prevalence and are less likely to take anti-diabetic medications than non-Hispanic White individuals [13]. Despite these disparities, AAPI are underrepresented in both epidemiological and genomic T2D research and the inherent diversity of these populations is often overlooked [14]. To our knowledge, this is one of the first times, across any disease outcome, that new methodologies for prediction are being developed with a focus on AAPI populations first, rather than as an afterthought.

\section{METHODS}
Our study population consisted of 65 participants. These participants were recruited directly from a Stanford Healthcare clinic. Participants were primarily aged 65–69 (28.6\%), 70–74 (25.4\%), and 75–79 (27.0\%), with smaller proportions aged 80–84 (9.5\%) and 85–89 (9.5\%), and 95–99 (1.6\%). 46.2\% of participants identify as Asian American and Pacific Islander, 32.3\% as White, 12.3\% as Hispanic, 7.7\% as Black, and 6.2\% as Middle Eastern/North African. The gender distribution of the population includes 70.5\% identifying as male, 29.5\% as female. The data for the project was obtained under IRB approval from Stanford University. 

We conducted unstructured interviews asking general conversational questions about patients’ life stories and experiences with diabetes, enabling participants to discuss topics they find significant. This approach ensures visibility into their lived experiences and highlights self-identified challenges and priorities. Interview lengths varied, with a minimum word count of 236, a maximum of 7,366, and most under 1,500 words. Alongside these narratives, we collected the full electronic health record  clinical data for each participant. The mean most recent A1C level among participants was 6.83, with a median of 6.6, a standard deviation of 1.05, and a range from 4.5 to 10.3.

We employ Stanford SecureGPT [15], which enables secure utilization of ChatGPT models for high-risk, protected health information. Using this platform, we applied LLMs to our dataset through three methods:
\begin{enumerate}
        \item \textbf{Extracting structured, free-form SDOH data}: Converting unstructured narrative content into structured, actionable insights.
        \item \textbf{Using extracted SDOH data for risk prediction}: Generating quantitative variables from qualitative narratives to serve as inputs for machine learning-based risk prediction models.
        \item \textbf{Predicting level of diabetes control}: Asking LLMs to predict a patient's level of diabetes control from the unstructured narratives.
\end{enumerate}

\subsection{Extracting structured, free-form SDOH data}

Because the interviews were unstructured, they covered a wide range of topics. To identify and organize themes within the dataset, we prompted ChatGPT-4o to read through each interview transcript, identify main themes and sub-themes, and provide three supportive quotes for each theme (Prompt 1 in Appendix). To further consolidate the dataset, we used sentence transformers to generate embeddings for the codes and applied agglomerative clustering to group similar codes. These clustered codes were manually reviewed and consolidated into representative codes, resulting in 16 key risk factors: level of knowledge about T2D management, social support, outlook towards their condition, and their day-to-day habits, socioeconomic status, education, occupation, immigration status, smoking status, alcohol consumption, physical activity, diet, air quality, social support, and access to green space, public transport, health services. 

Next, we aimed to systematically identify the presence of these risk factors across all interviews. We used retrieval-augmented generation (RAG) as our approach, leveraging its proven ability to reduce hallucinations and improve factual accuracy in responses [17]. First, we generated embeddings for each interview using OpenAI’s text-embedding-3-large embedding model [18]. Using the embeddings, our workflow retrieved the most relevant excerpts for each risk factor and passed the excerpts to ChatGPT-4o, which was prompted to provide bullet-point summaries of each patient’s relationship to the risk factor, along with supporting quotes from the transcript (Prompt 2 in Appendix). From this process, we identified five topics, each with three sub-topics, as the most frequently discussed across all interviews:

\begin{enumerate}
        \item \textbf{Socioeconomic Status}: Income Level, Housing, Financial Stress
        \item \textbf{Diet}: Diet Type, Food Preferences, Dietary Restrictions
        \item \textbf{Social Support}: Family Support, Friends, Social Networks
        \item \textbf{Health Services}: Healthcare Utilization, Satisfaction with Services, Barriers to Care
        \item \textbf{Information on Diabetes Management}: Knowledge Level, Self-Care Practices, Medication Adherence
\end{enumerate}

For each of these five topics and their subtopics, we developed a 1-to-5 rating scale (in Appendix), where higher scores indicated more positive outcomes or stronger alignment with best practices. The LLM was prompted to assign a rating based on the scale, provide a justification in concise bullet points, and include supporting quotes from the transcript (Prompt 3 in Appendix). If the topic was not in the interview, the LLM was asked to provide a rating of -1. 

To validate these ratings and ensure their accuracy, we manually reviewed five interviews, comparing the LLM-generated ratings and justifications against human interpretations of the interviews.

\subsection{Using extracted SDOH data for risk prediction}
 In this section, we build risk models for predicting a patient's A1C level using three feature sets of data from the 65 patients:
 \begin{itemize}
    \item \textbf{Social Determinants of Health (SDOH)}: We use the numerical ratings of each patient’s SDOH risk factors, extracted by ChatGPT-4o from the interviews as described in Method A. 
    \item \textbf{Patient laboratory results}: For each patient, we collect their most recent values for triglycerides, high-density lipoprotein (HDL),  low-density lipoprotein (LDL), glucose, and creatinine. This serve as a standard control dataset for the risk prediction task, as prior studies have demonstrated that glucose, lipid panels, and creatinine are key predictors of diabetes onset, progression, and related complications [19, 20, 21, 22]. 
    \item \textbf{Combined SDOH + labs}: We combine both datasets to simulate the most realistic use case, where SDOH data augments existing information in electronic health records to improve predictive performance.

\end{itemize}
 
We considered aiming to predict hospitalization risk based on frequency of ED-to-hospitalizations and ED visits, however, given the dataset size, the hospitalization data was sparse (73.9\% of patients did not have hospitalizations since the study start date in 2022). 

Prior to model development, the SDOH dataset underwent several preprocessing steps to ensure quality and robustness of the predictive models. Given that the interviews are free-form and cover a wide range of topics, not all themes necessarily overlap across patient narratives. This is expected and naturally results in missing entries in the extracted SDOH factors for some interviews. Approximately 17\% of the SDOH factors were missing across all patients and factors. Because our dataset is small, removing these missing values would significantly reduce the available data. We use K-Nearest Neighbors to impute the missing missing entries based on the similarity between observations. Then, to mitigate the impact of outliers and ensure that the features contributed comparably to the model, we scale the input data to a range of 0–1 to allow for predictable regularization behavior as well as preventing any sensitivity to the scale of inputs from the models. 

Each feature set was randomly divided into training and testing sets with an 80/20 split. The training set was used for model development and hyperparameter tuning. The testing set was reserved for evaluating the model performance.

For all feature sets, we explored linear and tree-based regression approaches to predict A1C levels, modeled as a continuous target variable. The models developed include: 
\begin{itemize}
    \item \textbf{Random Forest Regressor}: A tree-based ensemble model known for handling both linear and nonlinear relationships and high-dimensional effectively.
    \item \textbf{XGBoost Regressor}: Another tree-based ensemble model that optimizes for prediction accuracy through iterative refinement.
    \item \textbf{Lasso Regression}: A linear regression model with $L1$ regularization to encourage sparsity in feature selection through mitigation of overfitting.
    \item \textbf{Ridge Regression}: A linear regression model with $L2$ regularization to also prevent overfitting and better handle potential multicollinearity.
\end{itemize}

Linear regression was chosen for its effectiveness and simplicity in regression tasks involving a single dependent variable (A1C levels), while tree-based regression models were employed for feature importance analysis. Feature importance helps identify which variables most strongly influence model predictions. It is calculated using the Gini Importance metric, which measures the total reduction in error or impurity (such as mean squared error) attributed to each feature across all trees in the ensemble.

Model performance was assessed using $R^2$ score as the primary metric. The $R^{2}$ score measures how well the model fits the data, in other words, the proportion of variance in the dependent variable (A1C levels) that is predictable from the independent variables (SDOH factors and/ or lab data). For the tree-based models (Random Forest and XGBoost), hyperparameter tuning was performed using grid search with cross-validation, optimizing for the $R^{2}$ score. 

\subsection{Predicting level of diabetes control}
Unlike traditional regression models, which rely solely on structured numerical inputs as training data, LLMs are built on a foundation of extensive pretraining on diverse textual data and can extract and synthesize information from unstructured data. We tested the ability for OpenAI's ChatGPT-4o, o1, and o1-mini models, as well as DeepSeek's r1 model to predict the A1C level of the patient directly from patient interview. We first asked ChatGPT-4o to remove all mentions of A1C from the interview by replacing A1C numerical values with "[REMOVED]" (Prompt 4 in Appendix) in order in order to prevent models from relying on explicit A1C information and instead assess their ability to infer control from the interview content. The models were then prompted with the interview transcript and tasked with determining the A1C level of the patient and providing a justification with relevant quotes (Prompt 4 in Appendix). 

We categorized the A1C predictions and the patient's true A1C values into three control levels (low, medium, high) based on the distribution of the patient's true A1C values as the following:
\begin{itemize}
    \item \textbf{Low}: A1C $< 6.0 $ (16.9\% of patients)
    \item \textbf{Medium}: $6.0 \le$ A1C $\le 7.5$ (66.2\% of patients)
    \item \textbf{High}: A1C $> 7.5$ (16.9\% of patients)
\end{itemize}

To evaluate the accuracy of each model, we compared the control level of the model's prediction with the true control level of the patient. 

\section{RESULTS AND DISCUSSION}
 
\subsection{Extracting structured, free-form SDOH data}

Figure 1 illustrates the coverage of each subtopic across interviews. Among the most frequently discussed topics, income level, family support, social networks, and medication adherence were present in 90\% or more of the interviews. 

\begin{figure}[thpb]
      \centering
      \includegraphics[scale=0.3]{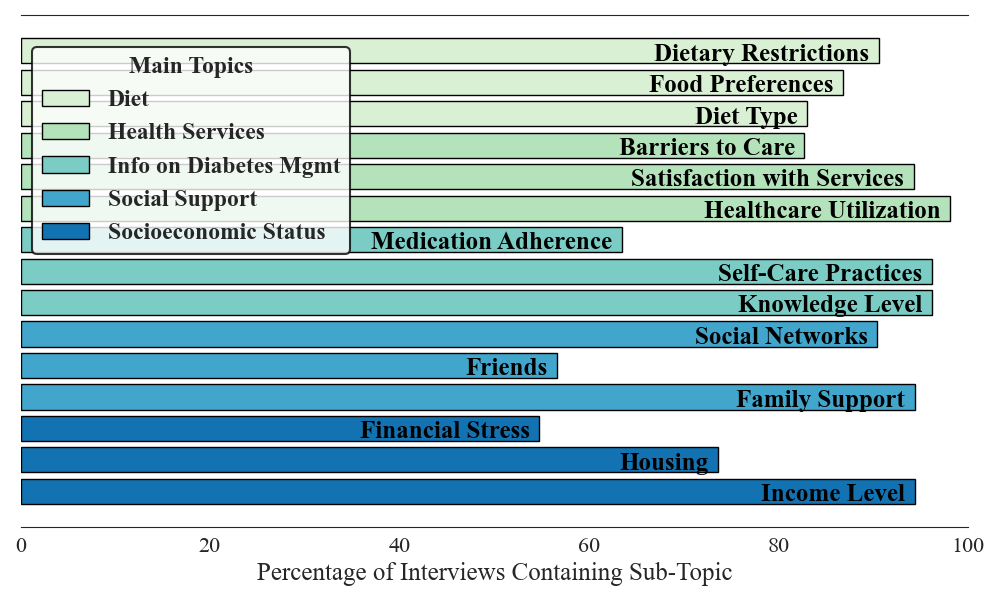}
      \caption{Percentage distribution of how frequently a specific topic was discussed across patient interviews.}
      \label{figurelabel}
   \end{figure}

Through qualitative analysis, we found that most summaries were relevant to the main topic. When analyzing subtopics within the same domain, we observed that the LLM  sometimes repeated information across similar subtopics (e.g., Dietary Restrictions and Diet Type). Separating subtopics allowed us to guide the LLM to retrieve more comprehensive and specific insights for each area, though at the cost of some duplication. Figure 2 is an example of a qualitative summary for one patient's "Social Support - Social Networks" factor.

 By asking the model to include quotes, we are able to verify the accuracy of extracted information and reduce the likelihood of hallucination by the LLM. This method is particularly valuable in clinical settings, where healthcare providers often lack time to deeply explore patients’ social determinants. Through our method, providers can quickly understand aspects of a patient's social context that influence their health. Furthermore, by being able to quantify a patient’s risk related to an SDOH factor, we can develop population-level data for risk modeling and public health studies, as demonstrated in Method B. 

\begin{figure}[thpb]
      \centering
      \framebox{\parbox{3in}{Rating: 4

Justification: The interviewee describes having a strong network of friends with whom they have consistent and meaningful interactions, indicating a high level of social support.

- The interviewee has a strong connection with long-time friends who provide consistent social interaction.

  [Quote]: "I can reach out to any of my friends, and we talk on the phone for an hour, whatever” 
  
- The interviewee values their friendships despite differing political views, indicating a deep and accepting relationship.

  [Quote]: "I've known her for 46 years. I know exactly what she's going to say. I know how her feelings are, and you know about this and that, and she's a conservative Republican, but I love her anyway."

- The interviewee frequently invites friends to enjoy their garden, showing an active effort to maintain and share their interests with friends.

  [Quote]: "I always invite my friends to my garden, because I can't go out and work in it anymore."

- Despite enjoying solitude, the interviewee emphasizes that they are never lonely due to their strong social connections and personal interests.

  [Quote]: "I like to be in my silence and and alone. And so people sometimes say to me, oh, you know, I don't want you to be lonely. It's never lonely. I am alone, but I'm never lonely, and I have my books, I have my artwork around me."
}}

      \caption{LLM generated summary for "Social Support - Social Networks" analysis for one patient}
      \label{figurelabel}
   \end{figure}

\subsection{Using extracted SDOH data for risk prediction} 
\begin{figure}[ht]
\centering
\resizebox{\linewidth}{!}{%
\begin{tabular}{lcccccc}
\toprule
 & \multicolumn{2}{c}{\textbf{SDOH}} & \multicolumn{2}{c}{\textbf{Labs}} & \multicolumn{2}{c}{\textbf{Combined}} \\
\cmidrule(lr){2-3}\cmidrule(lr){4-5}\cmidrule(lr){6-7}
\textbf{Model} & Train & CV & Train & CV & Train & CV \\
\midrule
Lasso          & 0.000 & \phantom{-}--0.038 & 0.001 & --0.008 & 0.001 & --0.090 \\
Ridge          & 0.248 & \phantom{-}0.046 & 0.028 & --0.027 & 0.217 & --0.015 \\
Random Forest  & 0.704 & --0.053 & 0.713 & --0.063 & 0.782 & \phantom{-}0.013 \\
XGBoost        & 0.995 & --0.194 & 0.948 & --0.112 & 0.999 & --0.103 \\
\bottomrule
\end{tabular}%
}
\caption{Training and cross‐validated $R^{2}$ by model and feature set.}
\label{fig:r2_model_performance}
\end{figure}

Figure 3 presents the performance of the four models in predicting A1C levels. Among them, only Ridge Regression on the SDOH-only dataset ($R^2 = 0.046$) and Random Forest on the combined dataset ($R^2 = 0.013$) achieved positive cross-validated $R^2$ values. A positive $R^2$ indicates that the model explains more variance in A1C levels than a baseline model that predicts the mean, reflecting modest predictive value on unseen data.

The low $R^2$ values, along with the disparity between the training and cross-validated $R^2$ values, suggest that the models are overfitting to the training data, likely due to the small sample size relative to the number of predictors. With relatively few observations and many variables, the models suffer from low degrees of freedom, making it difficult to reliably estimate residual variance and resulting in large standard errors in linear regression. 

When we augmented the SDOH-only models with traditional blood biomarker predictors, model performance further deteriorated, highlighting the challenges of high-dimensional inference in small samples. 
Since the blood biomarkers within the lab data are highly correlated both with one another and with the SDOH scores, introducing them inflates variance in parameter estimates and amplifies measurement error. Although ensemble tree methods are designed to reduce overfitting, they also tend to struggle with small datasets, producing overly specific splits that fail to generalize. In our case, this manifested in models whose Residual Sum of Squares (RSS) exceeded the Total Sum of Squares (TSS), i.e. worse than predicting the mean for every observation. These findings suggest that, without a substantially larger sample size or careful dimensionality reduction, adding more health-related variables may worsen rather than improve model validity.

\begin{figure}[ht]
\centering
\resizebox{\linewidth}{!}{%
\begin{tabular}{r l c}
\toprule
\textbf{Rank} & \textbf{Feature} & \textbf{Importance} \\
\midrule
1  & Glucose                                          & 0.216 \\
2  & Creatinine                                       & 0.187 \\
3  & Triglyceride                                     & 0.124 \\
4  & Socioeconomic Status -- Financial Stress         & 0.098 \\
5  & Diet -- Diet Type                                & 0.064 \\
6  & HDL                                              & 0.056 \\
7  & LDL                                              & 0.050 \\
8  & Diet -- Food Preferences                         & 0.047 \\
9  & Social Support -- Friends                        & 0.032 \\
10 & Social Support -- Family Support                 & 0.027 \\
\bottomrule
\end{tabular}%
}
\caption{Feature Importance for Combined Feature Set, Random-Forest}
\label{fig:feature_importance}
\end{figure}

 Figure 4 shows the feature importance analysis of the Random-Forest model on the combined dataset. Glucose, creatinine, and triglyceride levels have the most importance followed by Financial Stress and Diet Type in the combined model. This indicates that in combination, objective biomarkers remain powerful predictors, but are limited with a small sample size when assessing overall model prediction capability. Although biomarkers such as glucose, creatinine, and triglycerides are known to be correlated with A1C levels in the broader literature, our data-constrained study shows a collapse in their predictive power, likely due to the very few effective degrees of freedom for the models, given the large feature space in contrast to a small dataset. As a result, even though these features have higher importance than the SDOH features, they still aren't able to provide enough predictive value within our models, but still provide insight into the general standing of the different features within our model.

\subsection{Predicting level of diabetes control}
Our results indicate that the LLM can reasonably process and contextualize patient narratives. As shown in figure 5, the comparative evaluation demonstrates that GPT‑4o attains the highest overall accuracy (60\%), outperforming o1 (53.85\%), DeepSeek (50.77\%), and o1‑mini (50.82\%). 


 

\begin{figure}[ht]
\centering
  \resizebox{\linewidth}{!}{%
    \begin{tabular}{lcccc}
      \toprule
      \textbf{Model} & \textbf{Low} & \textbf{Medium} & \textbf{High} & \textbf{Overall Accuracy} \\
      \midrule
      GPT4‑O   & 0.0\% (0/11)   & 75.6\% (34/43) & 38.5\% (5/11) & \textbf{60.0\% (39/65)} \\
      O1   & 11.1\% (1/11)   & 68.2\% (30/43) & 33.3\% (4/11) & 53.8\% (35/65) \\
      O1‑Mini   & 0.0\% (0/10)   & 64.2\% (26/40) & 37.0\% (5/11) & 50.8\% (31/61) \\
      DeepSeek   & 0.0\% (0/11)   & 67.4\% (29/43) & 32.0\% (4/11) & 50.8\% (33/65) \\
      \bottomrule
    \end{tabular}%
      }
\caption{Categorical and overall accuracy for each model across 65 patients. o1-mini did not provide a response for 4 patients.}
\label{fig:a1c_model_accuracy}
\end{figure}

Through qualitatively analyzing A1C prediction justifications, we found that every justification included commentary on the person's lifestyle, whereas fewer mention strictly clinical markers. This pattern indicates that the models rely more on social- and behavioral-oriented cues when judging a patient’s level of diabetes control. This suggests that, even without prompting LLMs to extract SDOH factors as in Method 1, LLMs can be used to generate concise accounts of the patient’s behavioral and environmental contexts.

ChatGPT-4o has unique behaviors that could explain its performance advantage. ChatGPT‑4o justifications typically include exactly one quote about the patient's risk factors, lifestyle, and clinical experience. The quotes tend to be from three different parts of the transcript, whereas the other models tended to pick quotes that are from within the same section of the interview. This results in ChatGPT-4o having more balanced responses that potentially capture contrasting elements. In contrast, the reasoning models (o1, o1-mini, DeepSeek R1) have quotes that fit a more cohesive narrative. For example, for a medium-control patient, ChatGPT-4o uses the quotes \textit{“I walk the dog every morning”} and \textit{“doctor said pre‑diabetic two years ago,”} citing both improvement and historical warning to accurately determine medium control. For the same patient, o1 chooses \textit{"I walk the dog every morning"} and \textit{"cutting carbs"} which suggest lifestyle improvement and lead the model to determine low control. Additionally, ChatGPT-4o is more literal. ChatGPT-4o takes “no difficulty” and “controlled” statements at face value, whereas o1 cross‑checks self‑report with context (e.g., continued medication). ChatGPT-4o also tends to be more conservative in risk weighting. For example, any unresolved risk (e.g., medication lapses, comorbidities) nudges ChatGPT‑4o into a higher A1C level, whereas o1 is more optimistic and, if a patient speaks about self-improvement, the prediction is a lower A1C level. 

None of the models handle the low category well. For ChatGPT-4o, all patients were incorrectly classified as medium likely due to mentions of medication use and past diagnosis of diabetes. The models performed equally well for the high category, although for different cases. We found that high‑A1C patients often discussed both past crises and recent lifestyle improvements or medication changes and, without temporal anchoring, models typically averaged the two narratives to determine medium.

While ChatGPT-4o demonstrated moderate accuracy overall, certain edge cases revealed scenarios where the model was "fooled" by the contextual complexity of patient narratives. These cases provide interesting insights into the model's strengths and limitations:
\begin{itemize}
    \item \textbf{Patient 1}: The model incorrectly classified this patient as having severe diabetes. This was likely due to the patient's mention of near-blindness, a symptom often associated with advanced-stage diabetes. However, their actual A1C level was low, suggesting that the blindness was unrelated to their diabetes condition.
    \item \textbf{Patient 2}: The model predicted a lower A1C level for this patient based on their statement, \textit{"I am prediabetic and shouldn’t be classified as diabetic."} In reality, their true A1C level was higher, indicating that the patient's self-reported status misled the model.
    \item  \textbf{Patient 3}: This patient had a low A1C level, which might suggest better diabetes control. However, their interview revealed that they had been hospitalized multiple times and had been on insulin therapy for over 20 years. This context indicates a much more severe diabetes condition than their A1C alone would suggest.
\end{itemize}

These findings highlight the potential and challenges of using LLMs for nuanced healthcare tasks, and the need for manual verification of the LLM's assessments using the quotes and justification provided by the LLM. 

\section{CONCLUSIONS}

This work serves as a proof of concept for integrating unstructured patient narratives into clinical workflows through the use of large language models (LLMs). We envision a workflow in which patients provide unstructured information about their lives through a variety of convenient options – pre-appointment interviews, free-form text boxes, virtual recordings, etc – and large language models (LLMs) generate actionable, verifiable summaries, quantitative risk scores, and assessments of health management. This approach highlights the potential for transforming how social determinants of health (SDOH) and other qualitative data are utilized in healthcare to provide more personalized and effective care. 

The main limitation of this study is the relatively small sample size of 65 participants, which affected the ability of machine learning models to find strong correlations or achieve robust predictive performance. Additionally, the reliance on self-reported narratives introduces the potential for social desirability bias, where patients might unintentionally report their experiences and behaviors in a more favorable light. We also compare LLM output to each patient’s most recent A1C level and attempt to predict that value using machine learning models. Incorporating longitudinal A1C data may offer a more meaningful comparison because it reflects trends and variability in control over time, aligning more closely with the narrative nature of the patient interviews.

In order to validate the accuracy compared to a physician, future work involves comparing A1C predictions and justifications with at least two physician experts. Such a comparison would provide deeper insights into the model’s accuracy and limitations. The reliance on Stanford SecureGPT, while ensuring secure handling of protected health information, also poses challenges related to accessibility and scalability. Locally hosted LLMs could address this issue, though they may introduce additional costs.

Further research is required to determine how these tools can be seamlessly integrated into clinical workflows. Understanding the practical requirements of physicians and tailoring the tool to meet their specific needs will be critical for maximizing its impact and adoption. Developing strategies to balance efficiency with transparency will be essential to building trust among clinicians and patients.

Through these advancements, we aim to move beyond a proof of concept towards a clinically validated, scalable system. Such a system has the potential to enhance diabetes management by incorporating key social and contextual factors into the clinical workflow and risk prediction models. We also hope to expand this approach to other chronic disease areas, leveraging the flexibility of general LLMs to address broader healthcare challenges.








\section*{APPENDIX}
\scriptsize
\subsection{LLM prompts}
\textbf{Prompt 1:}
You are an AI assistant specializing in qualitative analysis. Read the following interview transcript with a diabetes patient and identify the main themes and sub-themes discussed. For each theme, provide:
- A concise code (theme) that represents the key concept. 
- Up to three relevant direct quotes from the transcript that illustrate the theme. 
Provide the output in valid JSON format as a list of objects, where each object has the following structure:
{{
 "code": "Code representing the theme",
 "quotes": ["Quote 1", "Quote 2", "Quote 3"]
}}
Do not include any additional text or explanations outside the JSON.
Interview Transcript:
(interview text) 
Extracted Codes and Quotes:

\textbf{Prompt 2:}

You are an expert analyst extracting key information from an interview with a diabetes patient about their life story. The patients are all older than 50. **Task**: Extract and list only the key pieces of information specifically about "{topic}" from the following interview excerpts. 
**Instructions**: 
- Provide the information as a bullet-point list. 
- Start each bullet point with a standardized format: "- [Keyword][Time Frame]: Detail" 
- Use only the following predefined keywords for "\{topic\}": \{keywords\}
- Pay close attention to the time frames mentioned, and accurately reflect whether the information is about the past, present, or future.
- Indicate the time frame in each bullet point using [Past], [Present], or [Future]. 
- Focus exclusively on "\{topic\}" and exclude any unrelated details.
- If there is no information about "{topic}" in the interview excerpts, respond exactly with "No information available." 
- Do not include introductions, explanations, summaries, or conclusions. 
- Limit the list to a maximum of 5 bullet points. 
**Example**: 
- [Exercise][Past]: Used to play basketball in college. 
- [Diet][Present]: Currently avoids sweets and limits rice intake. 
- [Work Routine][Future]: Plans to retire next year. 
**Interview Excerpts**: context 
**Answer**: 

\textbf{Prompt 3:}
       Subtopic: \{subtopic\_label\} 
       Below is the rating scale for this subtopic:\{scale\_text\} 
       Here is the relevant content from the interview: 
       \{subtopic\_text\} 
       TASK: 
       1) Decide on an integer rating for the subtopic based on the scale above from 1..5, or -1 if not present. 
       2) Provide a short justification or rationale for why you chose that rating. 
       3) Then follow these instructions for summarizing and quoting material from the interview: 
       Provide a concise and specific answer in 3-5 bullet points. Each bullet point must include: 
   1. A clear summary of the insight or opinion in a single sentence. 
   2. A full supporting quote from the interviews, written verbatim, with no omissions or ellipses.  
   Ensure the quote is at least 2-3 sentences long and provides enough context for the reader to understand its meaning fully.
   3. If contradictory opinions exist, provide them in a separate bullet point with a full quote. 
   4. Ensure the entire output adheres to the above criteria.
   Formatting: <your summary> 
   Quote: <verbatim quote> 
       Return your answer in the following structure: 
       Rating: <numeric> 
       Justification: <1-2 sentence reason> 
       Summary And Quotes: <Bullet points following instructions> 

\textbf{Prompt 4:}
        Please remove all explicit mentions of A1C levels from the following interview text. Replace any specific A1C numbers or ranges with '[REMOVED]' while preserving the rest of the context and meaning of the sentences. Keep all other diabetes-related information intact.
        Interview Content: 
       \{interview\_text\} 
        Please return only the modified text without any additional commentary.

\textbf{Prompt 5:}
       Based on the following interview content, please: 
       1. Predict the person's A1C level to .1 accuracy and only use one number, not a range.
       2. Do not use mentions of A1C in the text to determine the level of diabetes control.
       3. Provide a detailed justification with relevant quotes from the interview 
       Interview Content: 
       \{interview\_text\} 
       Please format your response exactly as follows: 
        A1C Level: [A1C level prediction]
       Justification: [Your detailed analysis]
       Supporting Quotes: [At least 2-3 relevant quotes from the interview]
       
\subsection{Scale for Rating Patients on SDOH factors}
\textbf{Socioeconomic Status - Income Level}

\begin{itemize}
    \item \textbf{1:} Extreme financial hardship; below the poverty line; unable to meet basic needs (food, shelter).
    \item \textbf{2:} Low income; struggles with consistent basic needs but has some access to social support or assistance.
    \item \textbf{3:} Moderate income; meets basic needs but has no financial flexibility.
    \item \textbf{4:} Stable income; meets needs and has moderate financial security.
    \item \textbf{5:} High income; strong financial security with savings/investments.
    \item \textbf{-1:} Not mentioned or irrelevant in the interview.
\end{itemize}

\textbf{Socioeconomic Status - Housing}
\begin{itemize}
    \item \textbf{1:} Homeless or living in unstable, temporary housing.
    \item \textbf{2:} Inconsistent housing; risk of eviction or unsafe conditions.
    \item \textbf{3:} Stable housing but low-quality or unsafe environment.
    \item \textbf{4:} Stable, safe housing; moderate quality.
    \item \textbf{5:} High-quality, stable housing in a safe neighborhood.
    \item \textbf{-1:} Not mentioned or irrelevant in the interview.
\end{itemize}

\textbf{Socioeconomic Status - Financial Stress}
\begin{itemize}
    \item \textbf{1:} Severe financial stress; constant worry about affording essentials.
    \item \textbf{2:} Frequent financial stress; periodic struggle meeting expenses.
    \item \textbf{3:} Moderate financial stress; occasional worries but generally manageable.
    \item \textbf{4:} Minimal financial stress; able to pay bills comfortably most of the time.
    \item \textbf{5:} No financial stress at all; feels fully secure financially.
    \item \textbf{-1:} Not mentioned or irrelevant in the interview.
\end{itemize}

\textbf{Diet - Diet Type}
\begin{itemize}
    \item \textbf{1:} Diet dominated by unhealthy, processed foods.
    \item \textbf{2:} Primarily unhealthy with occasional healthy items.
    \item \textbf{3:} Balanced mix but with some indulgences.
    \item \textbf{4:} Mostly healthy with rare unhealthy moments.
    \item \textbf{5:} Consistently healthy, nutrient-rich diet.
    \item \textbf{-1:} Not mentioned or irrelevant in the interview.
\end{itemize}

\textbf{Diet - Food Preferences}
\begin{itemize}
    \item \textbf{1:} Strong preference for high-calorie, low-nutrient foods.
    \item \textbf{2:} Limited interest in healthy foods.
    \item \textbf{3:} Moderately open to healthy options.
    \item \textbf{4:} Prefer healthier foods, tries to avoid junk.
    \item \textbf{5:} Actively seeks out nutrient-rich, healthy foods.
    \item \textbf{-1:} Not mentioned or irrelevant in the interview.
\end{itemize}

\textbf{Diet - Dietary Restrictions}
\begin{itemize}
    \item \textbf{1:} Ignores recommended restrictions entirely.
    \item \textbf{2:} Very quick to lapse or cheat on them.
    \item \textbf{3:} Moderate adherence to dietary restrictions, but not perfectly consistent.
    \item \textbf{4:} Generally follows restrictions with occasional slips.
    \item \textbf{5:} Fully compliant; no major slips.
    \item \textbf{-1:} Not mentioned or irrelevant in the interview.
\end{itemize}

\textbf{Social Support - Family Support}
\begin{itemize}
    \item \textbf{1:} No family support; may feel isolated.
    \item \textbf{2:} Minimal or inconsistent family assistance.
    \item \textbf{3:} Some reliable help, but not comprehensive.
    \item \textbf{4:} Reliable and significant family support.
    \item \textbf{5:} Very strong or exceptional family network.
    \item \textbf{-1:} Not mentioned or irrelevant in the interview.
\end{itemize}

\textbf{Social Support - Friends}
\begin{itemize}
    \item \textbf{1:} No friends; isolated socially.
    \item \textbf{2:} Minimal friend support; rare interactions.
    \item \textbf{3:} Moderate friend support; some emotional or practical help.
    \item \textbf{4:} Strong friend network; consistent help.
    \item \textbf{5:} Exceptional friend support; highly dependable and engaged.
    \item \textbf{-1:} Not mentioned or irrelevant in the interview.
\end{itemize}

\textbf{Social Support - Social Networks}
\begin{itemize}
    \item \textbf{1:} Completely disconnected; no broader community ties.
    \item \textbf{2:} Limited social involvement; rare group activities.
    \item \textbf{3:} Some involvement with local community or social groups.
    \item \textbf{4:} Active in community or group settings regularly.
    \item \textbf{5:} Exceptionally integrated into multiple networks.
    \item \textbf{-1:} Not mentioned or irrelevant in the interview.
\end{itemize}

\textbf{Health Services - Healthcare Utilization}
\begin{itemize}
    \item \textbf{1:} No regular healthcare visits; major gaps in care.
    \item \textbf{2:} Rare or only emergency visits.
    \item \textbf{3:} Irregular but some routine check-ups.
    \item \textbf{4:} Consistent appointments and fairly comprehensive care.
    \item \textbf{5:} Excellent utilization; very proactive approach.
    \item \textbf{-1:} Not mentioned or irrelevant in the interview.
\end{itemize}

\textbf{Health Services - Satisfaction with Services}
\begin{itemize}
    \item \textbf{1:} Deeply dissatisfied; ongoing issues.
    \item \textbf{2:} Mostly dissatisfied; some minor positive points.
    \item \textbf{3:} Neutral or mixed feelings; basics met.
    \item \textbf{4:} Generally satisfied; no major issues.
    \item \textbf{5:} Extremely satisfied; fully trusts providers.
    \item \textbf{-1:} Not mentioned or irrelevant in the interview.
\end{itemize}

\textbf{Health Services - Barriers to Care}
\begin{itemize}
    \item \textbf{1:} Intense barriers (cost, transport, discrimination).
    \item \textbf{2:} Multiple major barriers limiting access.
    \item \textbf{3:} Some barriers but partially manageable.
    \item \textbf{4:} Few barriers; mostly minor.
    \item \textbf{5:} No barriers to care.
    \item \textbf{-1:} Not mentioned or irrelevant in the interview.
\end{itemize}

\textbf{Information on Diabetes Management - Knowledge Level}
\begin{itemize}
    \item \textbf{1:} No knowledge of diabetes management.
    \item \textbf{2:} Very little awareness of key concepts.
    \item \textbf{3:} Moderate knowledge; some gaps remain.
    \item \textbf{4:} Good knowledge base; fairly solid understanding.
    \item \textbf{5:} Excellent knowledge, possibly well-educated about diabetes.
    \item \textbf{-1:} Not mentioned or irrelevant in the interview.
\end{itemize}

\textbf{Information on Diabetes Management - Self-Care Practices}
\begin{itemize}
    \item \textbf{1:} No self-care for diabetes management.
    \item \textbf{2:} Rare or minimal attempts at self-management.
    \item \textbf{3:} Some consistent efforts but not comprehensive.
    \item \textbf{4:} Regular, reliable self-care habits.
    \item \textbf{5:} Exemplary self-care; very proactive and thorough.
    \item \textbf{-1:} Not mentioned or irrelevant in the interview.
\end{itemize}

\textbf{Information on Diabetes Management - Medication Adherence}
\begin{itemize}
    \item \textbf{1:} Never follows medication schedule.
    \item \textbf{2:} Very poor adherence; frequent lapses.
    \item \textbf{3:} Moderate adherence; occasional misses.
    \item \textbf{4:} Strong adherence; minor lapses only.
    \item \textbf{5:} Perfect adherence; no missed doses.
    \item \textbf{-1:} Not mentioned or irrelevant in the interview.
\end{itemize}

\end{document}